# Weakly Supervised Multi-Task Learning for Cell Detection and Segmentation

*Alireza Chamanzar[1,2], Yao Nie[1]*

[1] Roche Tissue Diagnostics, Digital Pathology, Santa Clara, CA, USA, [2] Carnegie Mellon University, Pittsburgh, PA, USA

**ABSTRACT**

Cell detection and segmentation is fundamental for all downstream analysis of digital pathology images. However, obtaining the pixel-level ground truth for single cell segmentation is extremely labor intensive. To overcome this challenge, we developed an end-to-end deep learning algorithm to perform both single cell detection and segmentation using only point labels. This is achieved through the combination of different task orientated point label encoding methods and a multi-task scheduler for training. We apply and validate our algorithm on PMS2 stained colon rectal cancer and tonsil tissue images. Compared to the state-of-the-art, our algorithm shows significant improvement in cell detection and segmentation without increasing the annotation efforts.

## 1. INTRODUCTION

Cell detection and segmentation is an important and fundamental step towards analysis of pathology tissue images. Cells' population, density, morphology and stain intensity metrics are examples of information which can be obtained from the detected and segmented cells in a tissue slide image for diagnosis purpose and/or stain quality measurement. In this paper, we propose a weakly supervised multi-task learning algorithm for automated end-to-end cell detection and instance segmentation.

Recent studies in cell segmentation based on deep learning methods have reported superior performance [1, 2, 3, 4, 5] compared to the traditional image processing methods such as Watershed [6]. Comparing to tasks such as classification and segmentation of easy-to-label objects (e.g., cars, trees, animals, etc.), cell detection and instance segmentation face additional challenges. For example, stained cells have large variation of contrast to the background; highly clustered cells have touching or overlapped boundaries and hard to segment individually; and most importantly, obtaining the pixel-level ground truth for cell instance segmentation is extremely labor intensive. In a recent work [1], Qu et al. proposed a weakly-supervised method for nuclei segmentation in H&E histopathology images, which requires only point label annotation and extracts pixel-level labels using Voronoi transformation and unsupervised pixel clustering. In this paper, we use a similar approach for nuclei stained cell detection and segmentation in immunohistochemistry (IHC) images. To better tackle the aforementioned challenges, we propose to treat cell detection and instance segmentation as separate tasks and train the model through multi-task training process. Meanwhile, we employ multiple point label encoding methods to generate task oriented pixel-level labels to facilitate the multi-task training. As a result, the point label encoding through local pixel clustering and repel codes [10] improve the detection and segmentation of weakly stained cells and highly clustered cells, respectively. For training, we explore the newly released Ranger optimizer [7], which is a combination of RAdam [8] and LookAhead [9] optimizers.

The details of the proposed algorithm are discussed in Section 2, followed by the performance results and comparison to the state-of-the-art algorithms in Section 3. Lastly, in Section 4, we discuss the limitations of our algorithm and suggest possible directions for improvements.

## 2. METHOD

In this section, we discuss the details of the dataset and the proposed algorithm, including preprocessing, task oriented point label encoding, architecture of the model and the multi-task training process.

### 2.1. Dataset

We used immunohistochemistry (IHC) PMS2 stained colorectal cancer and tonsil tissue slides. The dataset includes 256 512×512 images at the resolution of 0.5µm/pixel, covering tumor, peri-tumor, normal tissue, intra- and inter-follicular regions in the slides. This dataset has a rich variety of nuclei for the detection and segmentation tasks, e.g., positive cells (dark or weak brownish stains) and negative cells (bluish stains) in different shapes and sizes, with sparse or highly clustered spatial distribution.

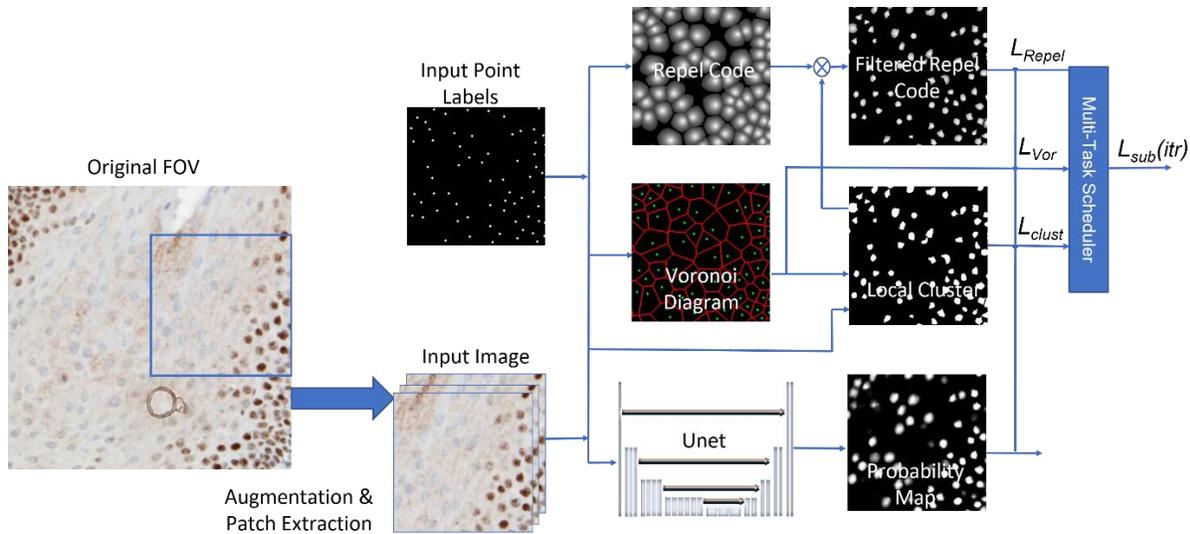

Figure 1: An overview of the proposed algorithm: pixel-level label extraction based on cell point labels and multi-task deep learning method using a loss scheduler and Unet model with a ResNet encoder.

## 2.2. Preprocessing and augmentation

We split the dataset into training (80%), validation (10%), and testing (10%) sets, making sure that each set has all types of tissue regions (i.e., tumor, peri-tumor, normal tissue, etc.). We extracted small patches of 250×250 pixels from the original images. To increase the size of the training set, data augmentation was performed, including horizontal and vertical flip, random resize, affine transform, rotation and crop. This resulted in a training set of ~3000 small images. As the last preprocessing step, we normalized the training set by mean subtraction and division by the standard deviation of the RGB channels separately. We applied the same normalization on the images in the validation and testing set.

## 2.3. Point label encoding and pixel-level label extraction

We design our algorithm in a weakly supervised fashion by applying three encoding methods on the cell point labels (see Fig. 1): (I) **Voronoi transformation** defines sub-regions consisting of all pixels closer to a particular point label than to any other. The sub-region partition lines are used as pixel-level label and help the highly clustered cells not to be merged together in segmentation. (II) **Local pixel clustering** is a new method we designed to extract pixel-level labels that help precise nuclei boundary delineation. Specifically, for each Voronoi sub-region defined in (I), k-means clustering algorithm is applied to extract the pixel-level labels of the nuclei and the background based on each pixel's RGB channel values concatenated with the distance of each pixel to the cell center point label. Using this local k-means, we can extract nuclei pixels which are located around the point labels and have local color contrast to the background. In comparison to the state-of-the-art in [1], where a global pixel clustering approach is used, our local clustering approach significantly improves the quality of the pixel-level labels especially for the weakly stained cells, as shown in Fig. 2. (III) **Repel encoding** is recently proposed as an enhanced center encoding for cells in [10], which defines a 2D decaying function with peak located at the cell center point label. Compared to the commonly used Gaussian and proximity encoding [10], the repel code decays faster for cells which have shorter distance to the neighboring cells. Therefore, we exploit the repel code in our algorithm to promote better cell separation in the segmentation task and at the same time better center localization for the detection task. In addition, to promote better nuclei boundary delineation, the extracted repel code is multiplied by the local pixel clustering label mask to ensure the background pixels to have zero value in the repel code map, which is referred to as the "filtered repel".

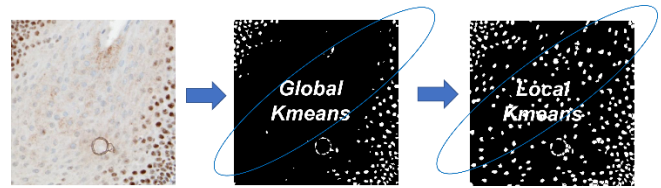

Figure 2: Performance of the proposed local clustering method vs. global clustering method in [1]. Weakly stained cells are well retained through the local clustering approach.

## 2.4. Model

We use Unet architecture [11], where the encoder layers are replaced with the convolutional layers of ResNet50, which is pre-trained on the ImageNet dataset [12, 13].

## 2.5 Multi-task scheduler

For training loss calculation, the nuclei and background probability maps, which are two output channels of the model are compared with the three extracted pixel-level labels using different loss functions. The cross entropy loss function is used for the binary labels (i.e., Voronoi and local pixel cluster labels) and the mean squared error (MSE) loss function is used for the repel code label, as defined in equations (1-3):

$$L_{clust}(t,o) = -\frac{1}{n*m}\Sigma_{i=1}^{n}\Sigma_{j=1}^{m}[t(i,j)\cdot \log o(i,j) + (1-t(i,j))\cdot \log(1-o(i,j))], \quad (1)$$

$$L_{vor}(t,o) = -\frac{1}{n*m-|ignored|}\Sigma_{\substack{i=1,\\i\neq ignored}}^{n}\Sigma_{\substack{j=1,\\j\neq ignored}}^{m}[t(i,j)\cdot \log o(i,j) + (1-t(i,j))\cdot \log(1-o(i,j))], \quad (2)$$

$$L_{repel}(t,o) = \frac{1}{n*m}\Sigma_{i=1}^{n}\Sigma_{j=1}^{m}\big(t(i,j)-o(i,j)\big)^2, \quad (3)$$

where $o$ is the model output probability map, and $t$ is the corresponding target, i.e., Voronoi, repel or local pixel cluster labels. The pixels in the *ignored* set in (2) are ignored in the Voronoi loss function (as illustrated by the black pixels inside the Voronoi sub-regions in Fig. 1), such that only the pixels indicated by the red lines (used as background) and green dots (used as foreground) are included.

Since we use three different losses to train a single model, we need a strategy to combine them for updating the weights of the model in each training iteration. In [1], a naïve summation of the losses is used to construct a total loss. However, for a multi-task learning problem, naïve summation may not be the optimal solution, as the nature of the tasks can be very different. To address this issue, we propose using a multi-task scheduler. Specifically, in each training iteration, only one of the three losses is used to update the model weights using the following rule: assume "$i$" is the index of the training iteration, the scheduler chooses the Voronoi loss if "$i \% 3 = 0$", choose the repel loss if "$i \% 3 = 1$", and choose the cluster loss if "$i \% 3 = 2$", i.e.,

$$L_{Sub}^{i}(t,o) = \mathbb{1}_{(\{i|i\,\%\,3\,=0\})}(i)\cdot L_{vor}^{i}(t,o) + \mathbb{1}_{(\{i|i\,\%\,3\,=1\})}(i)\cdot L_{repel}^{i}(t,o) + \mathbb{1}_{(\{i|i\,\%\,3\,=2\})}(i)\cdot L_{clust}^{i}(t,o), \quad (4)$$

where $L_{Sub}^{i}(t,o)$ is the selected loss at the $i^{th}$ training iteration, and $\mathbb{1}_{A}(i)$ is an indicator function which takes value 1 if $i \in A$, and 0 otherwise. Since we randomly shuffle the dataset before we extract the small batches for each epoch (see Section 3), each training image has a chance to contribute to all three losses/tasks. This multi-task scheduler shows a better performance for each individual task in comparison with the naïve summation of losses proposed in [1], as discussed in the next section. We generate binary segmentation masks using an *argmax* function applied on the output probability maps, which sets the pixel values to *one* where the probability value of nuclei is higher than background and *zero* otherwise. In addition, we detect the cells by finding the location of local maxima, with minimum distance of 2.5μm, in the nuclei output probability map using the maximum filter in [17].

## 3. RESULTS

In this section, we discuss the details of algorithm implementation, and report the performance results for the cell detection and instance segmentation tasks, along with the comparison with the state-of-the-art algorithm [1].

### 3.1 Implementation details

We trained the model in PyTorch [14], using batch size of 8, 150 epochs, and 60900 total training iterations. The newly released Ranger optimizer [7], which is reported to be more efficient in updating the parameters, improved the performance of training. We use learning rate of 0.001. For the repel code, we use $\alpha = 0.05$, and $r = 70$, following equation (4) in [10].

**Table 1.** Segmentation performance of the proposed algorithm in comparison with the state-of-the-art algorithms.

| Method | ACC | F1 | Dice | AJI |
|---|---|---|---|---|
| Method in [1], without CRF | 0.887 | 0.681 | 0.664 | 0.460 |
| Our algorithm with task uncertainty [15] | 0.915 | 0.731 | 0.735 | 0.532 |
| Our algorithm with multi-task scheduler | **0.929** | **0.791** | **0.784** | **0.599** |

**Table 2.** Detection performance of the proposed algorithm in comparison with the state-of-the-art algorithms.

| Method | Precision | Recall | CCC |
|---|---|---|---|
| Method in [1], without CRF | 0.874 | 0.936 | 0.015 |
| Our algorithm with task uncertainty [15] | 0.964 | 0.907 | 0.997 |
| Our algorithm with multi-task scheduler | **0.941** | **0.925** | **0.998** |

### 3.2 Performance

We evaluate the performance of the proposed algorithm based on a test set with **~25k** cells. For the segmentation task, our algorithm with multi-task scheduler has a pixel-level accuracy of **92.9%**, pixel-level F1 score of **79.1%**, object-level Dice score of **0.784**, and the object-level Aggregated Jacard Index (AJI) score of **0.599**. For the detection task, the detection precision is **94.1%**, the detection recall is **92.5%**, and the concordance correlation coefficient [16] (CCC, $\alpha = 0.05$) of the detected cell count is **0.998**. The segmentation performance metrics are used as defined in [1]. The detection precision and recall metrics are defined as *TP/(TP+FP)* and $TP/(TP+FN)$, respectively, where *TP, FP,* and *FN* are the numbers of true detection, false detection, and missing detection. Compared to [1], the proposed algorithm shows significant improvement in cell segmentation and detection, without increasing the annotation efforts. We excluded the conditional random field (CRF) post-processing step, which is used in [1], in our comparison. According to [1], CRF doesn't perform well for highly clustered nuclei, which have high prevalence in our IHC dataset. We have also compared the performance of our multi-task scheduler method with the multi-task learning method based on the task uncertainty as proposed in [15], where an additional loss layer with learnable weights are defined to combine the different losses. The results show that the multi-task scheduler method achieves better performance compared to the task uncertainty method. The results of segmentation and detection are summarized in Table 1 and Table 2. The algorithm in [1] shows a very small CCC value in the detection task (see Table 1), which is due to the large number of local minima which are detected based on the nuclei probability map. Since the repel encoding is not used in [1], the segmented nuclei regions are mainly flat rather than peaky, which causes this low CCC. As a visual illustration of instance segmentation performance of our algorithm, in comparison with the proposed algorithm in [1], we include some sample test images and the segmentation mask overlaid on the original images as shown in Fig. 3.

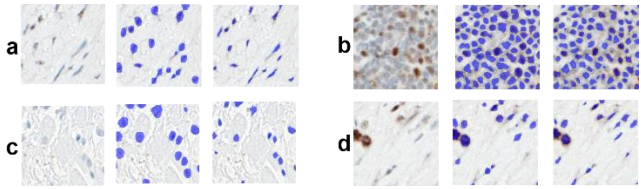

Figure 3: Segmentation performance illustration with sample test images (left image), the overlaid segmentation masks based on the algorithm proposed in [1] without CRF (middle image), and our algorithm (right image): (a) normal tissue region with elongated cells, (b) intra-follicular region with highly clustered cells, (c) normal tissue region with unstained cells, and (d) tumor region with weakly and strongly stained

## 4. DISCUSSION AND CONCLUSION

In this paper we propose an end-to-end cell detection and instance segmentation algorithm based on a multi-task learning approach. The three main challenges commonly encountered in cell detection/segmentation problems, i.e., lack of pixel-level ground truth, difficulty in detecting/segmenting weakly stained cells and highly clustered cells, are effectively addressed using the proposed algorithm. One of the limitations of this algorithm is the lower convergence speed using the multi-task scheduler method, which is the training expense we pay to gain performance efficiency during multi-task inferencing. In addition, the proposed method is most suitable for cells with distinguishable nuclei, for which the assumption of a convex shape and consistent stain color within each nucleus need be generally valid. On the other hand, by providing point labels for different cell types, the proposed method can be naturally extended to further address cell subtype classification problem without changing the proposed multi-task learning framework.

## 5. ACKNOWLEDGMENT

We thank Roche tissue early development and research group for providing us the PMS2 IHC dataset. We also thank our team members Andrew Deng, Nazim Shaikh and Kien Nguyen for helpful discussions.